\documentclass[letterpaper]{article} 
\usepackage{aaai25}  
\usepackage{times}  
\usepackage{helvet}  
\usepackage{courier}  
\usepackage[hyphens]{url}  
\usepackage{graphicx} 
\usepackage{natbib}  
\usepackage{caption} 
\usepackage{algorithm}
\usepackage{algorithmic}
\usepackage{threeparttable}
\usepackage{amsmath}
\usepackage{booktabs}
\usepackage{subcaption}
\usepackage{newfloat}
\usepackage{listings}
\usepackage{enumitem}
\usepackage{multirow}
\usepackage{xcolor}
\usepackage{soul}
\usepackage{animate}

\def\eg{{e.g.}}

\DeclareCaptionStyle{ruled}{labelfont=normalfont,labelsep=colon,strut=off} 
\floatstyle{ruled}
\newfloat{listing}{tb}{lst}{}
\floatname{listing}{Listing}
\title{PersonaMagic: Stage-Regulated High-Fidelity Face Customization\\ with Tandem Equilibrium}
\author{
    Xinzhe Li\textsuperscript{\rm 1}, Jiahui Zhan\textsuperscript{\rm 1,2}\thanks{Equal contribution.},
    Shengfeng He\textsuperscript{\rm 3}, Yangyang Xu\textsuperscript{\rm 4}, Junyu Dong\textsuperscript{\rm 1}, Huaidong Zhang\textsuperscript{\rm 5},\\
    Yong Du\textsuperscript{\rm 1}\thanks{Corresponding author.}\\   
}
\affiliations{
    \textsuperscript{\rm 1}Ocean University of China, China\\

    \textsuperscript{\rm 2}Shanghai Jiao Tong University, China\\
    \textsuperscript{\rm 3}Singapore Management University, Singapore\\
    \textsuperscript{\rm 4}Harbin Institute of Technology (Shenzhen), China\\
    \textsuperscript{\rm 5}South China University of Technology, China\\
    \{lixinzhe, zhanjiahui\}@stu.ouc.edu.cn, shengfenghe@smu.edu.sg, cnnlstm@gmail.com, huaidongz@scut.edu.cn, \{dongjunyu, csyongdu\}@ouc.edu.cn
}

\usepackage{bibentry}
\begin{document}

\maketitle

\begin{abstract}
Personalized image generation has made significant strides in adapting content to novel concepts. However, a persistent challenge remains: balancing the accurate reconstruction of unseen concepts with the need for editability according to the prompt, especially when dealing with the complex nuances of facial features. In this study, we delve into the temporal dynamics of the text-to-image conditioning process, emphasizing the crucial role of stage partitioning in introducing new concepts. We present PersonaMagic, a stage-regulated generative technique designed for high-fidelity face customization. Using a simple MLP network, our method learns a series of embeddings within a specific timestep interval to capture face concepts. Additionally, we develop a Tandem Equilibrium mechanism that adjusts self-attention responses in the text encoder, balancing text description and identity preservation, improving both areas. Extensive experiments confirm the superiority of PersonaMagic over state-of-the-art methods in both qualitative and quantitative evaluations. Moreover, its robustness and flexibility are validated in non-facial domains, and it can also serve as a valuable plug-in for enhancing the performance of pretrained personalization models.
\end{abstract}

\section{Introduction}
As a natural extension of research on controllability in diffusion models~\cite{ramesh2022hierarchical, rombach2022high}, personalized text-to-image generation~\cite{gal2022image, ruiz2023dreambooth} has emerged as a prominent task. By providing multiple images of a specific subject, users can introduce new concepts to a pre-trained text-to-image diffusion model, enabling it to synthesize the same subject in various contexts. However, despite recent advancements, current methods struggle to align generated outputs with users' envisioned images, particularly in face customization. The primary challenge lies in preserving the identity of a given face. While it is relatively straightforward to assess the consistency of generated results for common subjects (\eg, man-made objects, animals) based on contours and textures, achieving similar consistency in personalizing human faces, with their intricate features, is a more complex task.

\begin{figure*}
\centering
\includegraphics[width=\linewidth]{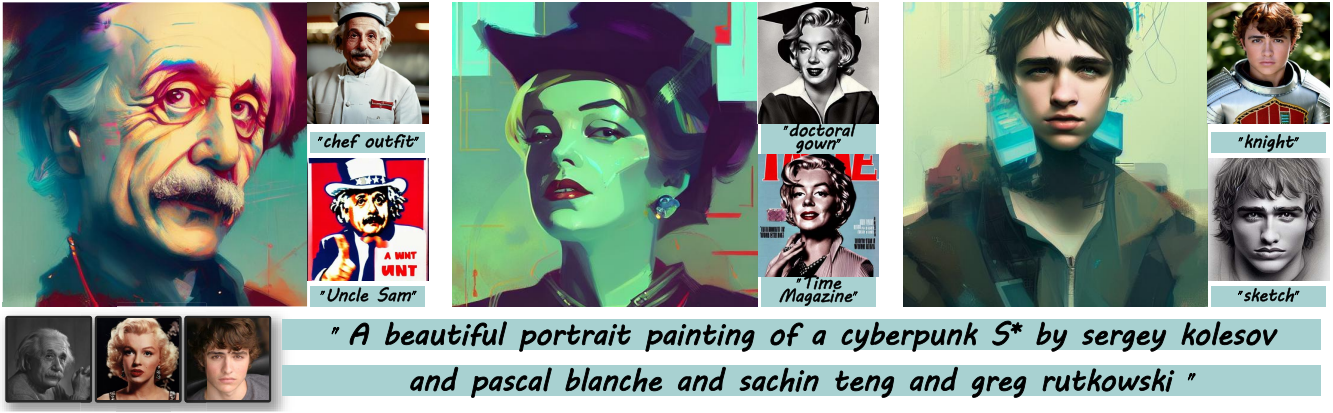}
\caption{\textit{PersonaMagic} seamlessly generates images of new roles, styles, or scenes based on a user-provided portrait. By learning stage-regulated embeddings through a Tandem Equilibrium strategy, our method accurately captures and represents unseen concepts, faithfully creating personas aligned with the provided prompts while minimizing identity distortion.}
\label{fig:teaser}
\end{figure*}

New concepts are typically represented as word embeddings~\cite{gal2022image} and integrated into prompts for customization. When training on a single image, the learned embedding often represents the entire image including background details rather than focusing on the target concept. As a result, personalized face generation usually requires 3 to 5 images of an individual in various scenes and poses to guide the cross-attention maps derived from the learned embedding toward the facial region. Several approaches~\cite{kumari2023multi, tewel2023key} attempt to fine-tune cross-attention layer parameters to adjust the attention of new concepts, but this requires substantial memory costs during training and introduces language drift~\cite{lee2019countering, lu2020countering}. Leveraging human-centric datasets, recent methods~\cite{ye2023ip, li2024photomaker} have trained personalization models to generate customized results with accurate identities and more natural human poses. However, we observe a decline in identity preservation when generating images of individuals not included in the training set.

In this paper, we address the fidelity-editability challenge using only a single image. Our main idea is first rooted in the understanding that consistent textual conditioning manifests varying control capabilities across different temporal phases, as reflected in the cross-attention maps. During an intermediate stage of the denoising process, the learned embedding tends to focus on the facial region rather than the entire image, even when training on a single image. Additionally, as noted in previous research~\cite{alaluf2023neural}, word embeddings capture different concept details across timesteps. This insight inspired us to develop dynamic embeddings within a specific timestep interval to achieve personalized face generation. During training, we divide the reverse process of diffusion models into dynamic and static stages, based on changes in cross-attention maps over time. In the dynamic stage, we introduce a lightweight network to acquire dynamic embeddings at different timesteps, effectively capturing user-provided face information. Conversely, the static stage employs fixed word embeddings corresponding to supercategory to stabilize training.

With the diffusion model frozen, the learned embeddings become the sole variable for minimizing denoising loss during training. This focus on learned embeddings may cause the diffusion model to overlook the semantics other than new concept. We observe that these neglected word embeddings have lower attention weights in the cross-attention layers of U-Net, resulting from the disproportionately high attention given to the learned embeddings in the text encoder. To address this, we introduce a Tandem Equilibrium (TE) strategy. During training, we input diverse text prompts into the text encoder and balance the attention weights of the new concept with other tokens, ensuring complete semantic representation. Unlike prior methods that required multiple images to generalize the expressiveness of the learned embedding across different scenarios, our TE strategy achieves this by directly operating within the text encoder, eliminating the need to pass latent image features to U-Net. This approach allows us to generate identity-preserving and semantically complete results with a single image, as shown in Fig.~\ref{fig:teaser}.

In summary, the contributions of this work are threefold:
\begin{itemize}
	\setlength{\itemsep}{0pt}
	\setlength{\parsep}{0pt}
	\setlength{\parskip}{0pt}
	\item We propose learning dynamic embeddings within a specific range to enable high-fidelity personalized face generation. Even from a single image, the learned embeddings generate the desired cross-attention maps, effectively preserving identity while improving efficiency.
	\item We introduce a TE strategy to regulate self-attention maps in the text encoder, ensuring that personalized results align closely with textual descriptions without the need for multiple images.
	\item Extensive quantitative and qualitative experiments validate our effectiveness, demonstrating a strong balance between textual alignment and identity preservation.
\end{itemize}

\section{Related Work}
\noindent\textbf{Text-to-Image Diffusion Models.} Diffusion models~\cite{song2019generative,song2020score,tang2022improved} have recently excelled in generating images from text~\cite{ramesh2021zero,gu2022vector,yu2022scaling,saharia2022photorealistic}. Noteworthy examples include GLIDE~\cite{nichol2021glide}, which crafts high-resolution images using diverse diffusion models, and DALL·E 2~\cite{ramesh2022hierarchical}, generating CLIP~\cite{radford2021learning} image embeddings from text through a diffusion model. Imagen~\cite{saharia2022photorealistic} enriches semantic information with a pre-trained text encoder~\cite{raffel2020exploring}. Stable Diffusion~\cite{rombach2022high} proposes denoising in a low-dimensional latent space through an autoencoder~\cite{esser2021taming}. While these methods can generate images aligned with text prompts, customizing a specific subject remains challenging. Our aim is to train an efficient network to introduce concept information about an unseen face into a pre-trained text-to-image diffusion model, enabling face customization across various scenes or styles. 

\begin{figure}
	\centering
	\includegraphics[width=\linewidth]{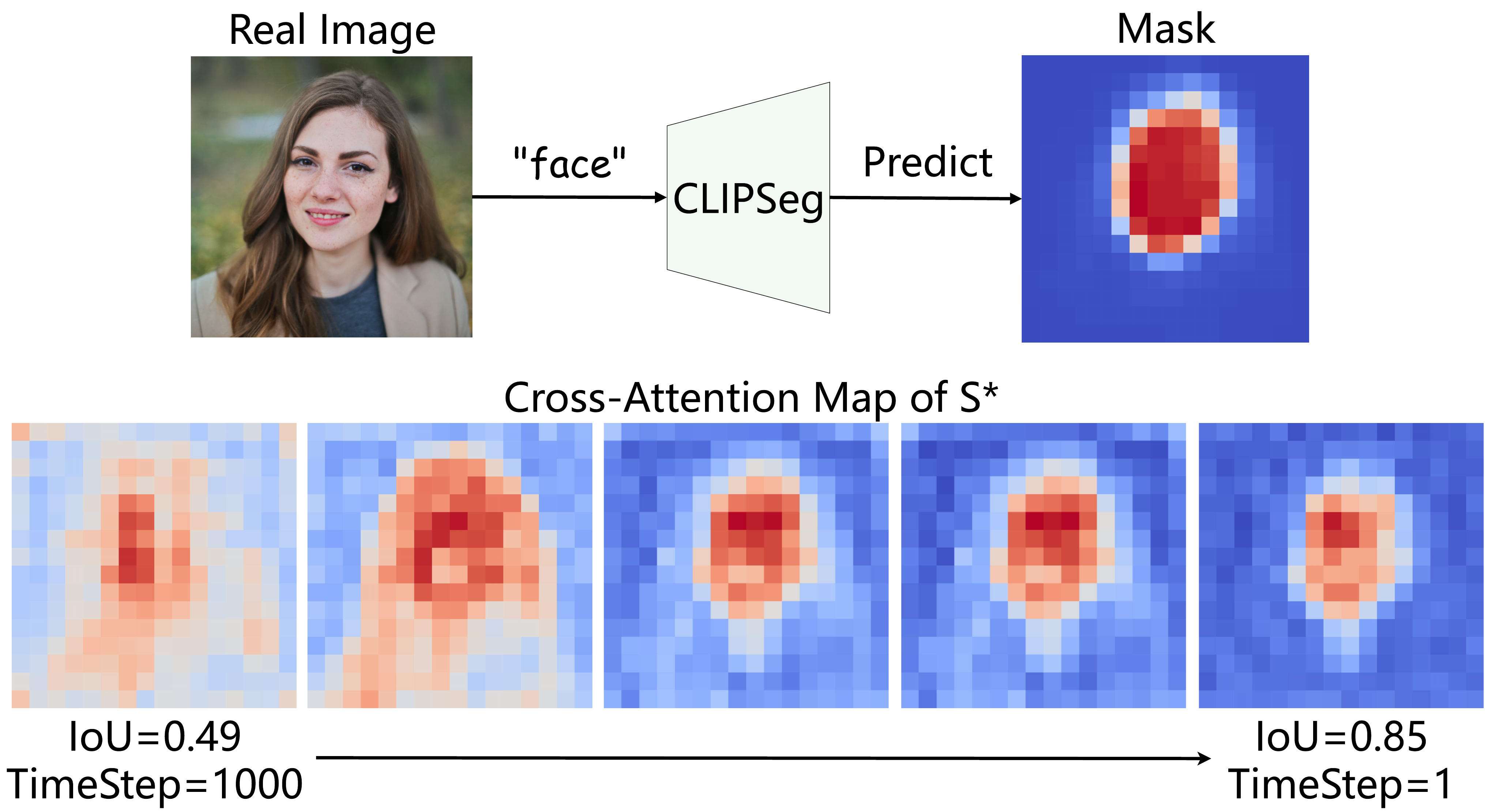}
	\caption{Cross-attention maps of $S*$ at each timestep. We calculate IoU with facial mask for stage partition. }
	\label{fig:att}
\end{figure}

\begin{figure*}
	\centering
	\includegraphics[width=\linewidth]{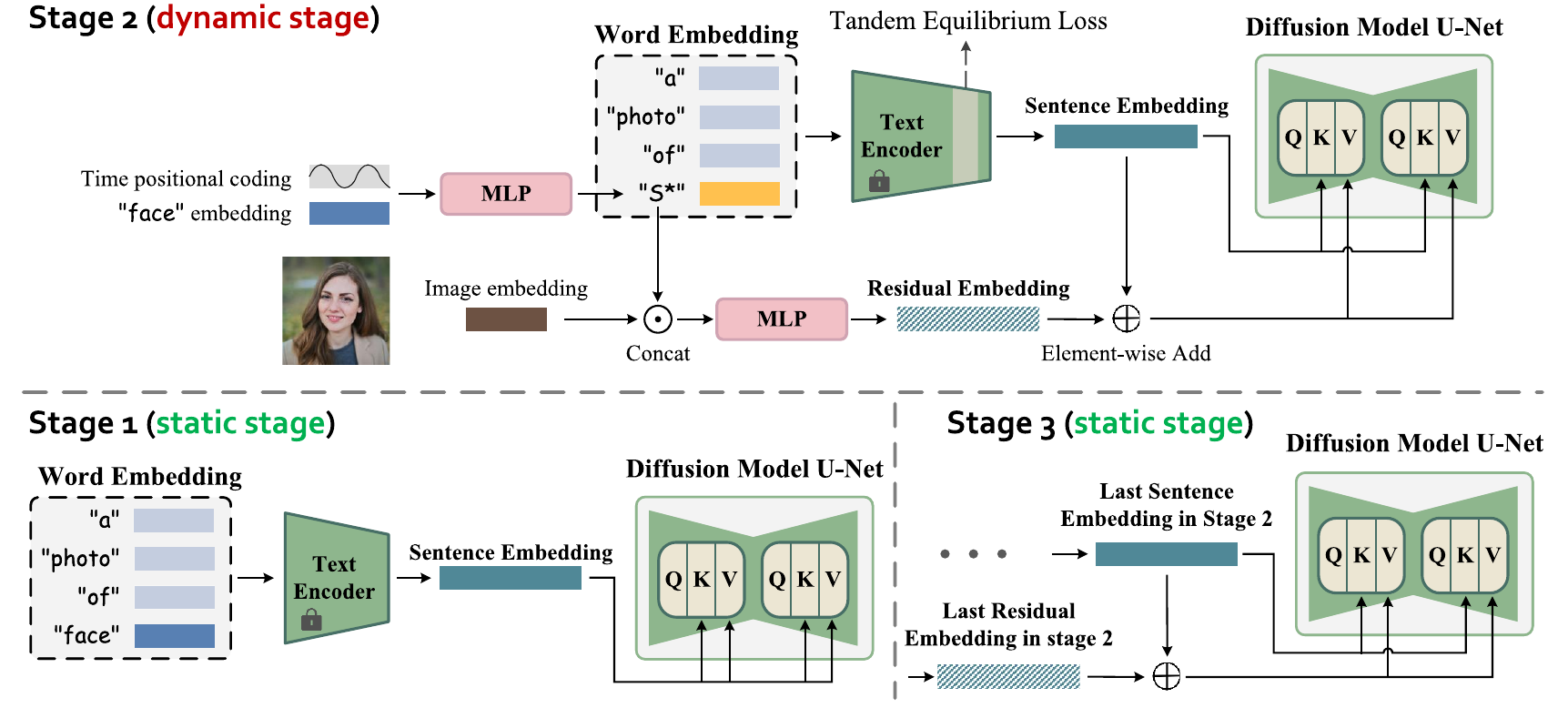}
	\caption{Overview of our pipeline. Given a single image, we learn a series of embeddings during dynamic stage to capture identity information effectively, while employing fixed embeddings in the static stages. The proposed TE strategy is applied in the text encoder, ensuring further alignment of personalized results with textual descriptions. 
	}	
	\label{fig:method}
\end{figure*}
\noindent\textbf{Personalized Image Generation.} Personalized generation methods find widespread applications across computer vision and graphics. Previous methods~\cite{yang2021discovering,xu2021continuity,nitzan2022mystyle, roich2022pivotal, alaluf2022hyperstyle,song2022editing} typically rely on GANs~\cite{karras2019style,karras2020analyzing,goodfellow2020generative}, encountering difficulties in handling out-of-domain images. Based on text-to-image diffusion models, Textual Inversion~\cite{gal2022image} proposed to optimize word embeddings for personalization. DreamBooth~\cite{ruiz2023dreambooth} recommends fine-tuning diffusion model parameters to introduce new concepts. However, these methods struggle to excel in both identity preservation and text similarity simultaneously. In contrast, we propose TE strategy, leading to trade-off in both aspects. 
Recently, several studies~\cite{ye2023ip,gal2023encoder,li2024photomaker} have trained on human-centric datasets, acquiring more comprehensive prior knowledge for personalization. For instance, IP-Adapter~\cite{ye2023ip} introduces a decoupled cross-attention strategy for semantic guidance. Photomaker~\cite{li2024photomaker} extracts ID embeddings from facial images to provide identity information to diffusion model. However, the performance of these approaches may be influenced by biased datasets, resulting in poor identity preservation for unseen faces. Instead, our method, guided by a self-contained stage-regulated text conditioning mechanism, efficiently customizes unseen persons with precise identity preservation. 

\section{Method}
\label{section:A}
\noindent\textbf{Stage-regulated Textual Conditioning. } 
We conduct an experiment to understand the temporal dynamics of the text-to-image conditioning process. Given a facial image, we use a rare token $S*$ to represent this new concept. Through a simple fully connected network and the textual condition ``A photo of $S*$'', we learn a set of word embeddings that vary across timesteps. We then visualize cross-attention maps of $S*$ at each timestep, as shown in Fig.~\ref{fig:att}. We observe that when timestep is large, cross-attention map of $S*$ is inaccurate, spreading its focus across the entire image. This indicates that the word embedding might be capturing background details, which is detrimental to personalized face generation. As the timestep decreases, cross-attention map of $S*$ progressively narrows down to a more precise facial region, suggesting that $S*$ can more effectively acquire accurate identity information at these timesteps. 

We believe that the reason for these phenomena is that in the early stage of the denoising process, noised image $X_t$ contains more noise, preventing $S*$ from accurately focusing on the facial region, which results in undesired cross-attention maps. Consequently, the learned embeddings struggle to capture useful concept information during this phase. As the noise in $X_t$ diminishes, the interference from redundant information decreases as well. However, prior research ~\cite{balaji2022ediffi} indicates that the control ability of text prompts over $X_t$ becomes weaker in the late stage of the denoising process. This suggests that learning word embeddings in the middle of the time schedule is a better choice than learning over the entire time schedule. 

Building on our analysis, we propose partitioning the time schedule in the reverse process of diffusion models into three intervals based on cross-attention maps. However, variations in facial regions across images affect the identity information captured by the learned embeddings, resulting in differences in the selected intervals. To reasonably define these stages, we initially train across the entire time schedule. After several iterations, the learned embedding, while not yet capturing accurate identity information, begins to show significant differences in cross-attention maps at different timesteps, without exhibiting overfitting. We utilize an existing semantic segmentation model, CLIPSeg~\cite{lueddecke22_cvpr}, to extract facial mask and calculated IoU with cross-attention maps. As shown in Fig.~\ref{fig:att}, we observe that IoU gradually increases as timestep progresses. Based on this, we set two thresholds, $\lambda_1$ and $\lambda_2$ ($\lambda_1 < \lambda_2$), to update the training intervals. We designate the phase where the IoU falls below $\lambda_1$ as the first static stage. During this stage, we cease training and instead use the supercategory embedding (\eg, face) directly for inference. Once the IoU surpasses the threshold $\lambda_2$, the learned embedding accurately identifies the facial region, but this also indicates that the noised image contains minimal noise, reducing controllability over the diffusion model. Consequently, we define this as the second static stage, where training ceases, and a fixed word embedding is used during testing. 

We designate the remaining intermediate interval as the dynamic stage. In this phase, we utilize a lightweight network that takes as input a time embedding and a super-category word embedding to generate dynamic embeddings, as illustrated in Fig.~\ref{fig:method}. Similar to NeTI~\cite{alaluf2023neural}, we incorporate a residual embedding into the sentence embedding output from the text encoder to serve as $V$ for the cross-attention layer, providing additional information that the text encoder alone cannot capture. Notably, we introduce the CLIP image embedding of the training image into the network to generate the residual embedding. This is because features extracted from images are beneficial for learning conceptual information, a finding supported by existing research~\cite{ye2023ip,li2024photomaker}.

\begin{figure}
	\centering
	\includegraphics[width=\linewidth]{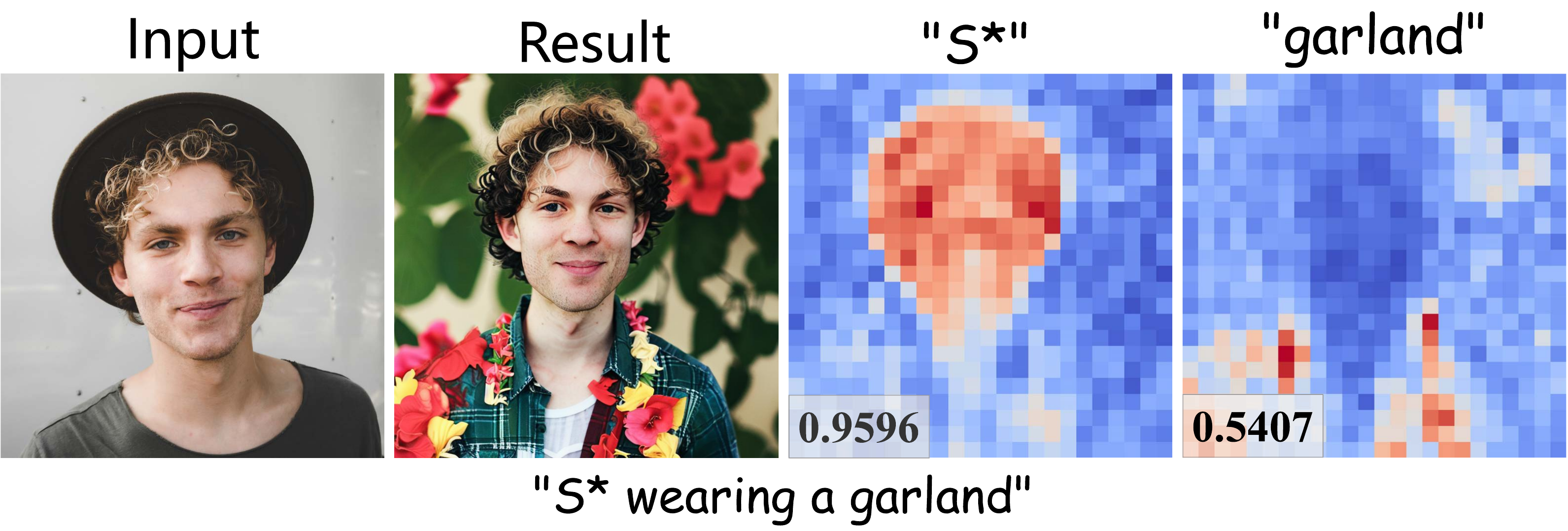}
	\caption{The overlooked semantic yield suboptimal attention map. Attention weights are annotated in the lower left corner of cross-attention maps. }
	\label{fig:att_unbalance}
\end{figure}

\noindent\textbf{Face Customization With Tandem Equilibrium. } 
\begin{figure}
	\centering
	\includegraphics[width=\linewidth]{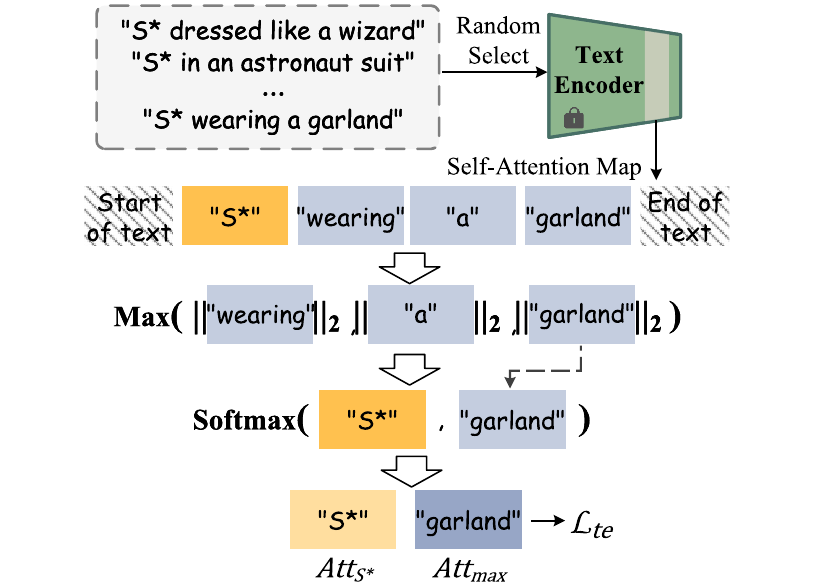}
	\caption{Illustration of the proposed Tandem Equilibrium. }
	\label{fig:att_balance_loss}
\end{figure}
The embeddings learned during the dynamic stage can more accurately focus on the facial region. This not only helps avoid overfitting but also improves training efficiency by narrowing the range of timesteps. However, we observe that even when $S*$ focus on the desired region, the diffusion model may still produce results inconsistent with the given text description under certain textual conditions. As shown in the first row of Fig.~\ref{fig:att_unbalance}, although the result accurately preserves the identity of the individual, it deviates significantly from the semantics of ``garland''. We use L2 norm as a global metric to measure the weight of each text token in the cross-attention map. The weight of $S*$ (0.9596) is notably higher than that of ``garland'' (0.5407), indicating that the diffusion model has neglected the semantic information of ``garland''. 

In U-Net, the cross-attention map is computed from $Q$ and $K$, where $Q$ is derived from the latent image features and $K$ from the sentence embedding. During training, we freeze diffusion model parameters to prevent language drift, making sentence embedding the sole variable affecting the cross-attention map. We further analyze the self-attention maps in the text encoder to understand this phenomenon. We find that in the shallow layers of text encoder, the attention weights for each text token are relatively uniform. As the layers deepen, nouns begin to dominate the attention, while prepositions have lower weights, due to the concrete semantic content typically associated with nouns. We believe that during training, in an effort to reduce denoising loss, text encoder overly emphasizes the semantics contained in the learned embedding while neglecting those in the original embedding. This is evident in the excessively high attention weights of $S*$, which may lead diffusion model to generate results that do not align with the given text description. 
\begin{figure*}
	\centering
	\includegraphics[width=\linewidth]{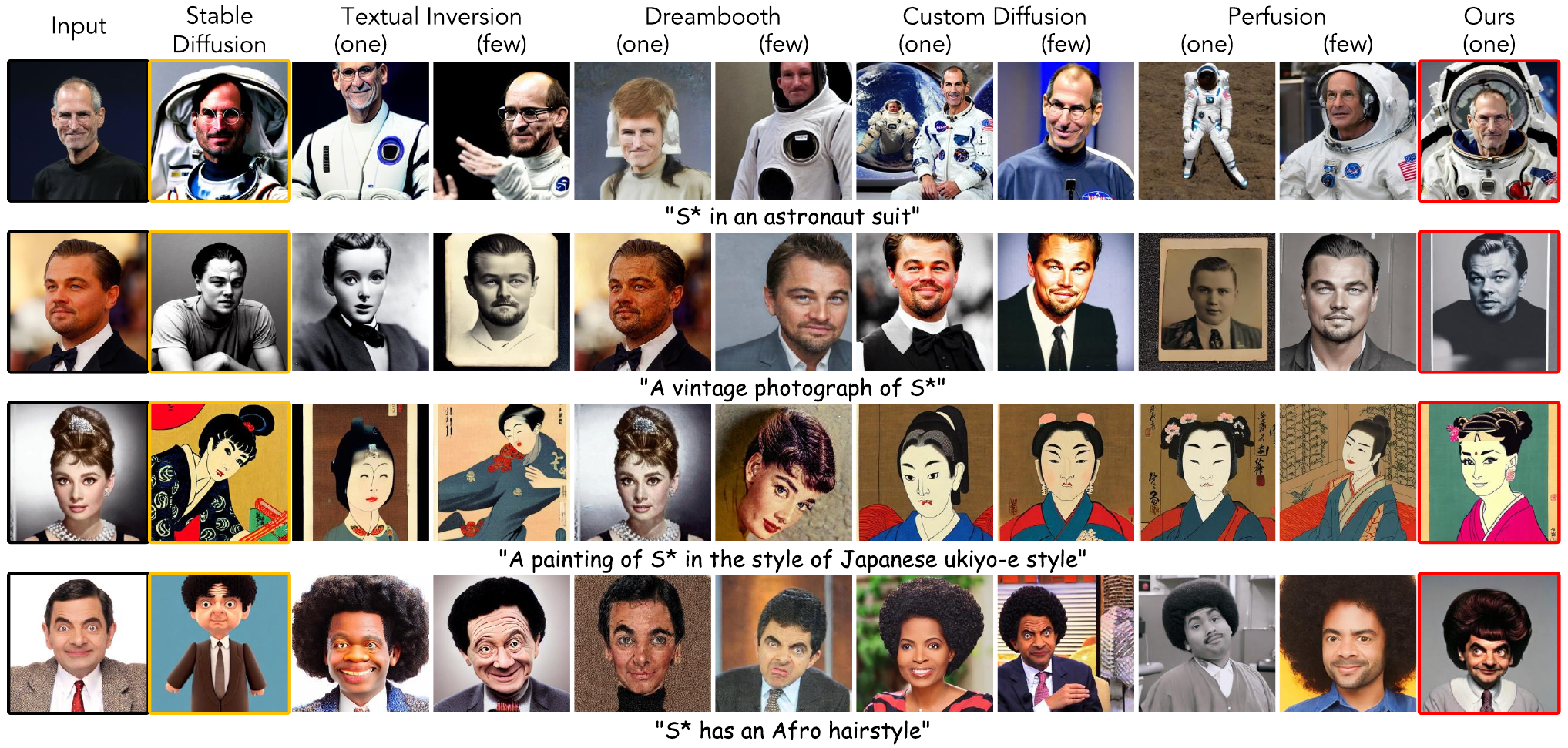}
	\caption{Qualitative comparison with state-of-the-art methods on celebrities.}
	\label{fig:comp_cele}
\end{figure*}

One straightforward solution is to minimize the L2 norm of the self-attention map corresponding to $S*$ within the text encoder. However, directly constraining the self-attention map can easily cause the diffusion model to overlook the semantics of $S*$, leading to results with inaccurate identities. Another solution is to manually set a threshold to constrain the attention weight of $S*$ within a specific range. However, since it varies across different text prompts, using a fixed threshold is not ideal. Therefore, the constraint should be adjusted dynamically based on text prompt. 

As shown in Fig.~\ref{fig:att_balance_loss}, we randomly input text prompt into the text encoder and extract self-attention maps in the final layer. Ignoring the ``start of text'' and ``end of text'' tokens, we identify the original embedding with the highest attention weight. We then applying a softmax function to its self-attention map with that of $S*$, obtaining $Att_{max}$ and $Att_{S*}$. We calculate tandem equilibrium loss $\mathcal{L}_{te}$ as follows: 
\begin{equation}
	\mathcal{L}_{te}=-\phi(Att_{S*}) \times \phi(Att_{max}),
\end{equation}
where $\phi(\cdot)$ denotes the summation function. Since the total sum remains constant, maximizing the product requires $\phi(Att_{S*})$ and $\phi(Att_{max})$ to be as close as possible. This ensures that the diffusion model balances its attention between $S*$ and other text tokens effectively.

\noindent\textbf{Loss Functions.}
We introduce a mask $M$ indicating the face region to calculate $\mathcal{L}_{mse}$, enforcing diffusion model to focus on denoising the masked region. It is formulated as: 
\begin{equation}
	\mathcal{L}_{mse}=||(\epsilon_\theta(X_t,t,y_t)-\epsilon)\cdot M||^2.
\end{equation}
To ensure the preservation of identity information from the given image $X_0$, we assess the similarity between the identity features of the noised image $X_t$ estimated to the clean image $X_{0|t}$ at time $t$ and $X_0$. Identity features are extracted using Arcface~\cite{deng2019arcface}, and the loss $\mathcal{L}_{id}$ is defined as follows:
\begin{equation}
	\mathcal{L}_{id}=1-\textrm{cos}(\textrm{Arcface}(X_{0|t}),\textrm{Arcface}(X_0))).
\end{equation}
As the diffusion model faces challenges in recovering an accurate clean image when $t$ is large, affecting the effectiveness of $\mathcal{L}_{id}$, we introduce a hyperparameter schedule, $\lambda_{id}(t)$:
\begin{equation}
	\lambda_{id}(t)=\textrm{cos}(\frac{t}{2T}\pi).
\end{equation}
Finally, our total objective $\mathcal{L}(t)$ is formulated as follows:
\begin{equation}
	\mathcal{L}(t)=\mathcal{L}_{te}+\mathcal{L}_{mse}+\lambda_{id}(t)\mathcal{L}_{id}.
\end{equation}
\begin{figure*}
	\centering
	\includegraphics[width=\linewidth]{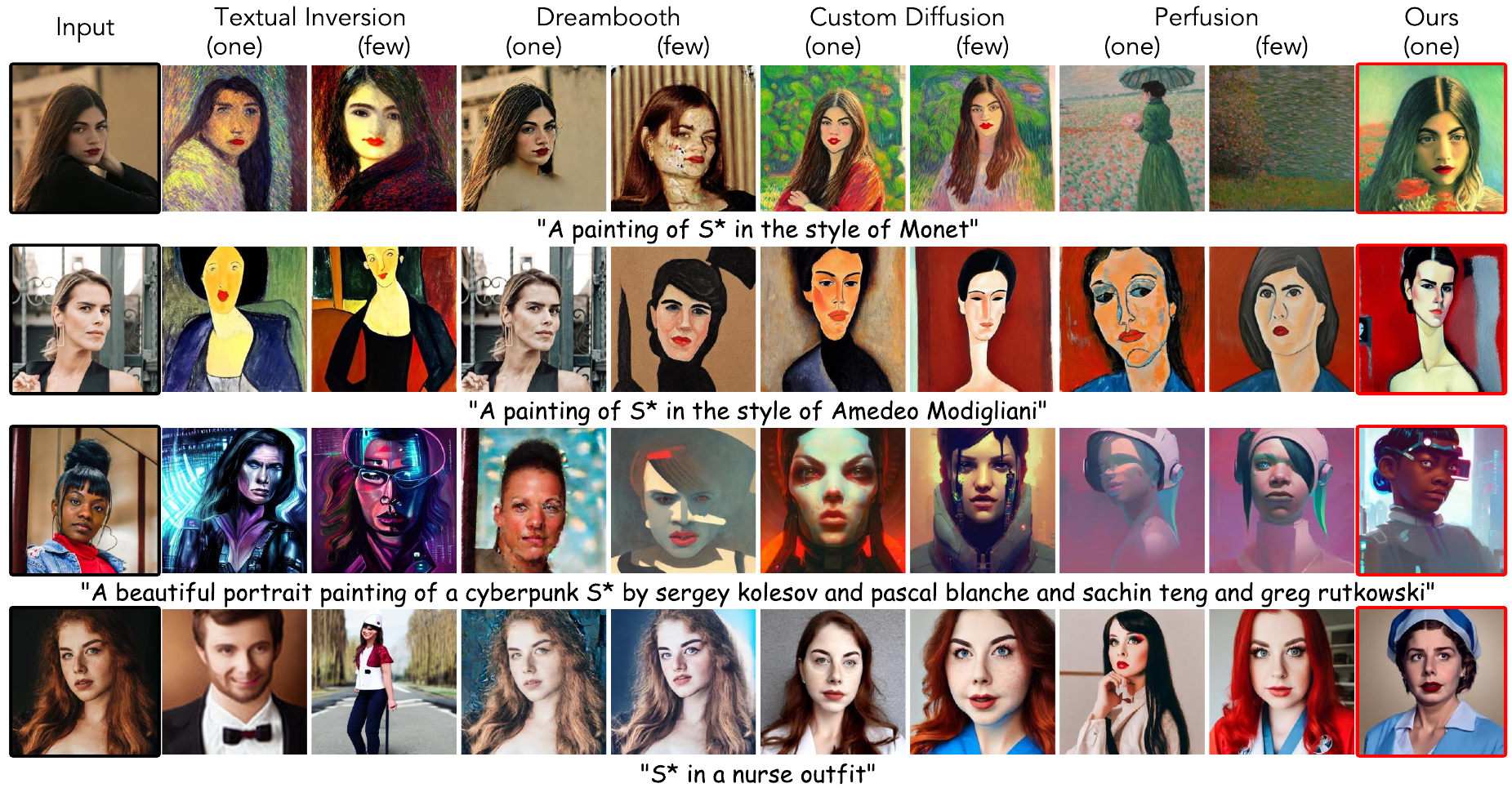}
	\caption{Qualitative comparison with state-of-the-art methods on non-celebrities.}
	\label{fig:comp_normal}
\end{figure*}

\section{Experiments}

\noindent\textbf{Competitors.}
We compare our method with several state-of-the-art personalization methods, including Textual Inversion~\cite{gal2022image}, DreamBooth~\cite{ruiz2023dreambooth}, Custom Diffusion~\cite{kumari2023multi}, NeTI~\cite{alaluf2023neural}, and Perfusion~\cite{tewel2023key}.
Since official implementations were unavailable for Perfusion and DreamBooth, we employed unofficial versions~\cite{von-platen-etal-2022-diffusers,Chen2023key} for our comparisons.
All comparative methods were conducted using their default settings.

\noindent\textbf{Metrics.}
We evaluate performance from two perspective to assess the effectiveness of proposed method.
For text similarity, we leverage the pre-trained CLIP~\cite{radford2021learning} model and calculate CLIPScore~\cite{hessel2021clipscore} between prompts and customized results, where the placeholder ``$S*$'' in prompts is substituted with ``face''.
For identity preservation, we use MTCNN~\cite{zhang2016joint} to detect faces in unaligned generated images and then apply CosFace~\cite{wang2018cosface} to evaluate identity similarity to the given faces. It is noteworthy that if no face is generated, the score is set to the minimum value of -1.

\noindent\textbf{Qualitative Evaluation.}
In order to visually demonstrate our generated effects, we collected some public images from the Internet, comprising some celebrity close-ups and portraits of non-celebrities, all of which are unaligned and encompass approximately 30 individuals.
We present generated outcomes under varies text prompts in Figs.~\ref{fig:comp_cele} and~\ref{fig:comp_normal}.
For each individual, we collected 3-5 images and randomly selected one as the input for one-shot setting. To ensure fairness, we also present the results of default few-shot setting for competitors.
Considering that all inputs in Fig.~\ref{fig:comp_cele} are celebrities, we also showcase the results of directly inputting prompts into pretrained diffusion model by replacing the placeholders ``$S*$" with the names of celebrities.
While generating directly from prompts leverages the extensive prior knowledge of Stable Diffusion to create celebrities with distinctive features, it may introduce biases from the given images, such as inconsistencies in Steve Jobs' hair or Leonardo DiCaprio's age.
It is evident that DreamBooth shows strong identity preservation, but its results often misalign with text prompts. This issue arises from training across the entire time schedule, leading to overfitting where cross-attention maps focus on the entire image. Contrastingly, our method confines training to the dynamic stage, directing embeddings to focus exclusively on facial regions.
Textual Inversion, Custom Diffusion, and Perfusion generate results that align with text prompts but lack accurate identity.
In contrast, our approach not only utilizes dynamic embeddings across multiple timesteps to convey intricate character information but also employs a lightweight network for learning, which better captures the correspondence between the embedding space and facial attributes.

\begin{figure}
	\centering
	\includegraphics[width=\linewidth]{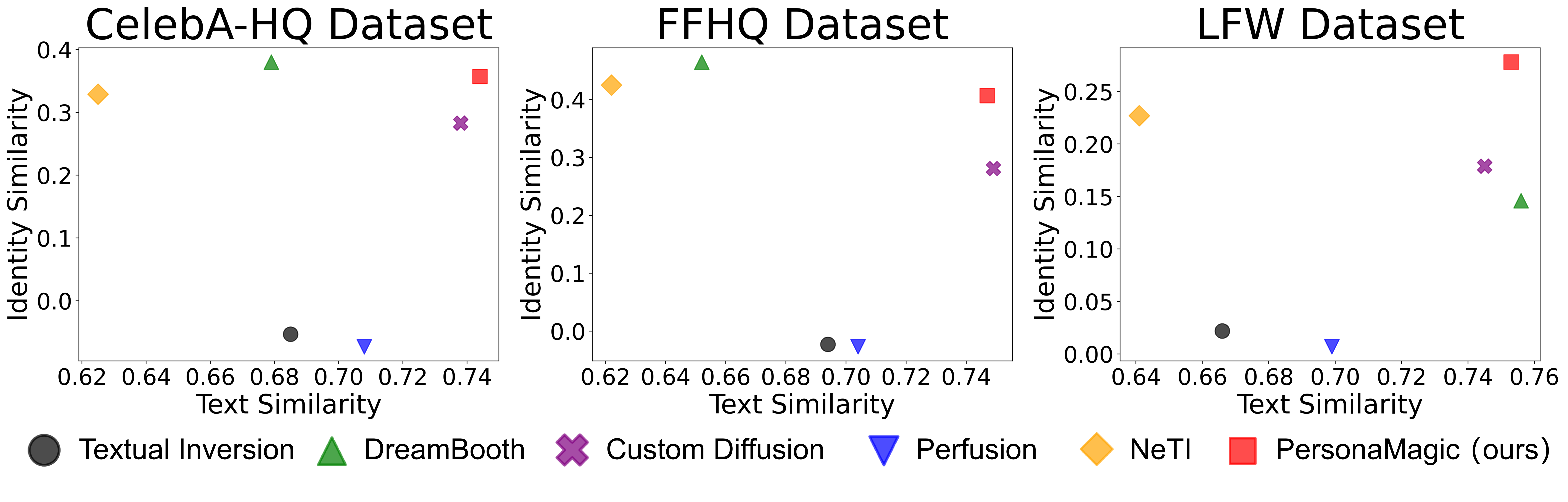}
	\caption{Quantitative evaluation on CelebA-HQ, FFHQ, and LFW datasets shows that our method sits on the Pareto front, highlighting its superiority over competitors. }
	\label{fig:Quantitative Evaluation}
\end{figure}

\noindent\textbf{Quantitative Evaluation.}
For quantitative evaluation of one-shot setting, we randomly selected 100 distinct images from CelebA-HQ~\cite{karras2017progressive} and FFHQ~\cite{karras2019style} datasets respectively and repeated five times for fair comparisons, gotten 500 images of each dataset. For few-shot setting, we use the LFW ~\cite{huang2008labeled} dataset, focusing on individuals that contain multiple images.
We collected 17 prompts involving diverse personalized modifications, with each individual and prompt inferred four times, resulting in 68 outcomes per person. We then calculated text similarity and identity preservation for each set, with the average score presented in Fig.~\ref{fig:Quantitative Evaluation}.
Our method outperforms competitors, sitting on the pareto front. Textual Inversion and Perfusion, struggling to capture intricate identity details, demonstrate low identity preservation. 
DreamBooth and NeTI excel in identity preservation but show lower text similarity. In contrast, we introduce the Tandem Equilibrium strategy to balance the semantic representation of text tokens, leading to better text similarity. Custom Diffusion balances concept fidelity and text alignment, but its performance remains below ours.

\begin{table}[t]
	\centering
	\setlength{\tabcolsep}{1mm}	
    \fontsize{9}{\baselineskip}\selectfont
	\begin{tabular}{l||c|c||c|c}
			\toprule
			\multirow{2}*{Method}  & \multicolumn{2}{c||}{CelebA-HQ}& \multicolumn{2}{c}{LFW} \\
			\cmidrule(lr){2-3}
			\cmidrule(lr){4-5}		
			&T-Sim. & I-Pre.&T-Sim. & I-Pre.\\					
			\midrule
			Custom Diffusion (CD) &0.737 &0.283 &0.744 & 0.179 \\
			\midrule
			Vanilla &0.610 &0.334 &0.631 & 0.226\\
			+Stage Regulation  &0.672 &\textbf{0.363} &0.665 & \textbf{0.289}\\
			+Tandem Equilibrium  &0.691 & 0.284 &0.668 & 0.206\\
			\midrule
			PersonaMagic &\textbf{0.744} &\underline{0.358} &\textbf{0.753} &\underline{0.278} \\
			\bottomrule
		\end{tabular}
		\caption{Ablation study on CelebA-HQ and LFW datasets. }
	\label{tab:Ablation Study}
\end{table}

\begin{table}[t]
	\centering
    \setlength{\tabcolsep}{1mm}	
    \fontsize{9}{\baselineskip}\selectfont
	\begin{tabular}{c|c|c|c|c|c|c}
			\toprule
			\multirow{2}*{Setting} & $\lambda_1$=0.5 &$\lambda_1$=0.6 & $\lambda_1$=0.7  & $\lambda_1$=0.8 &$\lambda_1$=0.7 &\textbf{$\lambda_1$=0.7} \\
			& $\lambda_2$=1.0 &$\lambda_2$=1.0 &$\lambda_2$=1.0 &$\lambda_2$=1.0 & $\lambda_2$=0.9 & \textbf{$\lambda_2$=0.8} \\
			\midrule
			T-Sim.&0.691&0.727&0.744&0.749&0.741 &0.744 \\
			\midrule
			I-Pre.&0.284&0.307&0.328&0.302&0.328 &0.358 \\
			\bottomrule
	\end{tabular}
	\caption{Evaluation of stage partition variants on CelebA-HQ. We set $\lambda_1=0.7$ and $\lambda_2=0.8$, as this configuration achieves the optimal balance between fidelity and editability. }
	\label{tab:stage partition}
\end{table}

\begin{figure}
	\centering
	\includegraphics[width=\linewidth]{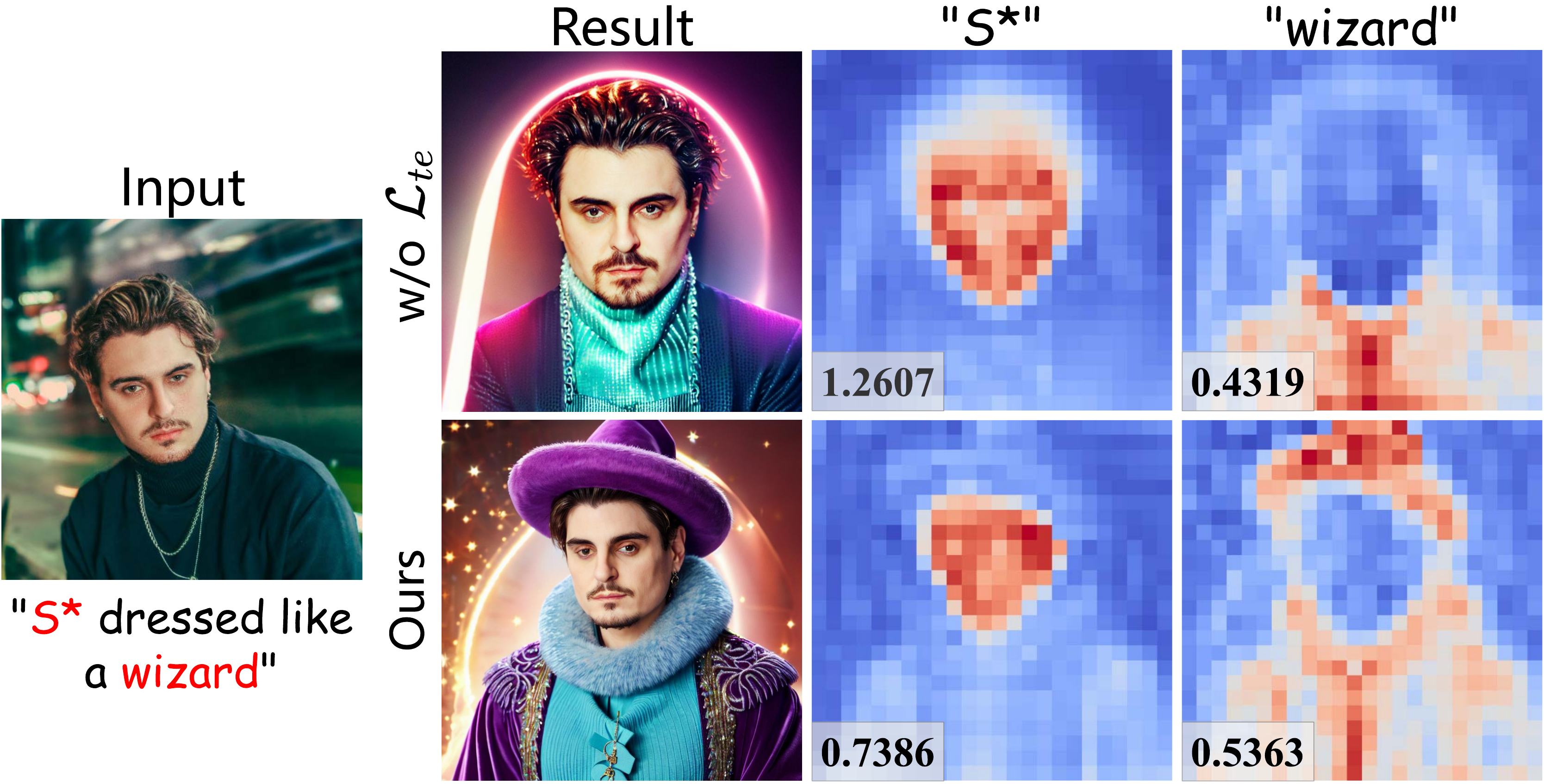}
	\caption{Customized results with and without $\mathcal{L}_{te}$ during training. Attention weights are annotated in the lower left corner of cross-attention maps. }
	\label{fig:att_balance_map}
\end{figure}

\noindent\textbf{Ablation Study.} We validated different components of PersonaMagic through ablation studies (Tab.~\ref{tab:Ablation Study}). First, we examined stage regulation's impact. In one-shot experiments on CelebA-HQ, our method improved identity preservation by 0.051 over the second best competitor CD due to learning a series of embeddings for new concepts, which outperforms optimizing a single embedding. However, text similarity decreased slightly due to the early use of learned embeddings, leading to misaligned spatial layouts with the prompt.

Stage regulation improved both text similarity and identity preservation, increasing text similarity from 0.610 to 0.672 by using supercategory embeddings in the initial static stage, reducing overfitting. The dynamic stage further enhanced ID preservation by focusing more on facial features.

To further explore stage regulation, we conducted a quantitative analysis (Tab.~\ref{tab:stage partition}). Initially, with $\lambda_1=0.5$ and $\lambda_2=1.0$, the denoising process was unpartitioned, using dynamic embeddings throughout, resulting in suboptimal performance. Setting $\lambda_1$ to 0.6 or 0.7 introduced a two-stage process: a static stage below $\lambda_1$ using supercategory embeddings and a dynamic stage above it with dynamic embeddings. Increasing $\lambda_1$ improved text similarity but reduced identity preservation at $\lambda_1 \geq 0.8$ due to limited dynamic stage timesteps. Based on the results, we chose $\lambda_1 = 0.7$. Further refinement of $\lambda_2$ divided the process into three stages, with a third static stage emerging as $\lambda_2$ decreased, using fixed embeddings. We observed that when $\lambda_2 < 0.8$, the dynamic stage became too constrained, impacting the capture of identity details. Therefore, we set $\lambda_1 = 0.7$ and $\lambda_2 = 0.8$. This strategy, based on IoU between cross-attention maps and real masks, generalizes well across datasets.

The TE strategy balanced text similarity and identity preservation by adjusting self-attention in the text encoder, aligning semantic strength between learned and original embeddings. Visualizations (Fig.~\ref{fig:att_balance_map}) showed TE reduced misalignment in cross-attention, improving personalized results.

Integrating both strategies, we surpassed CD in text similarity. Despite a slight reduction in identity preservation with TE, dynamic embeddings maintained accurate details, performing better CD. Few-shot experiments on LFW confirmed these strategies enhance performance, balancing text similarity and identity preservation.

\begin{table}[t]
	\centering
    \setlength{\tabcolsep}{1mm}		
    \fontsize{9}{\baselineskip}\selectfont
    \begin{tabular}{c||c|c||c|c}
        \toprule
        \multirow{2}*{Method} &\multirow{2}*{PhotoMaker} & PhotoMaker &\multirow{2}*{IP-Adapter} &IP-Adapter \\	
        & & w/ ours & & w/ ours\\	
        \midrule
        T-Sim.& 0.775 & \textbf{0.790} &0.762 & \textbf{0.778}\\	
        \midrule
        I-Pre.& 0.335 & \textbf{0.352} &0.348 &\textbf{0.359} \\
        \bottomrule
    \end{tabular}
	\caption{Universality of our method as a plug-in for pretrained personalization models. }
	\label{tab:Quantitative plugin}
\end{table}

\noindent\textbf{The flexibility of PersonaMagic.} Pretrained personalization models~\cite{li2024photomaker, ye2023ip} tend to perform suboptimally when applied to individuals outside the training dataset. This issue is not well-addressed by fine-tune based methods~\cite{ruiz2023dreambooth, kumari2023multi}, as it risks disrupting the pre-learned semantic knowledge in diffusion model. In contrast, our approach avoids this limitation by freezing model parameters during training, making it a flexible plug-in that can be integrated into pretrained face personalization models to enhance their performance.

To validate this, We integrated PersonaMagic into Photomaker~\cite{li2024photomaker} and IP-Adapter~\cite{ye2023ip} and conducted quantitative experiments on our collected images of non-celebrities, as shown in Tab.~\ref{tab:Quantitative plugin}. Introducing our method led to improvements in both text similarity and identity preservation for Photomaker and IP-Adapter. The increase in text similarity is largely attributed to our TE strategy, which guides the diffusion model to focus on overlooked semantics in the text prompt. Additionally, the Stage Regulation strategy enhances text similarity by helping Photomaker and IP-Adapter generate spatial layouts during early denoising that better align with the text description. This strategy also captures facial features that were previously missed, further improving identity preservation in personalized models. This conclusion is supported by visual comparisons, available in the supplementary materials.

\section{Conclusion}
In this paper, we present PersonaMagic, a high-fidelity face customization technique that utilizes a stage-regulated textual conditioning strategy based on a comprehensive analysis. We introduce a lightweight network to implement this conditioning mechanism through dynamic word embeddings, effectively capturing identity information while avoiding overfitting. Furthermore, we propose a tandem equilibrium loss to address the trade-off between text alignment and identity preservation. Extensive experiments demonstrate the superior performance of our method compared to state-of-the-art approaches, excelling in both fidelity and editability, and showcasing its effectiveness across various downstream customization tasks.

\section*{Acknowledgements}
This work is supported by the National Natural Science Foundation of China (No. 62102381, 41927805); Shandong Natural Science Foundation (No. ZR2021QF035); the National Key R\&D Program of China (No. 2022ZD0117201); the Guangdong Natural Science Funds for Distinguished Young Scholar (No. 2023B1515020097); the AI Singapore Programme under the National Research Foundation Singapore (Grant AISG3-GV-2023-011); and the Lee Kong Chian Fellowships.

\appendix
\section{Appendix}
\subsection{Implementation Details}
Our method is implemented using PyTorch 2.0.1, and the network architecture consists of two MLP modules, each containing two fully connected layers. One module generates dynamic embeddings, while the other produces residual embeddings. The outputs from these modules are processed through a LayerNorm~\cite{ba2016layer} layer, followed by a LeakyReLU activation function.

For training, we utilize the Adam optimizer~\cite{kingma2014adam} on a single NVIDIA RTX 3090 GPU. The learning rate is set to $5 \times 10^{-5}$, with a weight decay of 0.01. The hyperparameters $\beta_1$ and $\beta_2$ are set to 0.9 and 0.999, respectively. Training is conducted with a batch size of 2 for 1000 iterations. For stage partitioning, we set $\lambda_1$ to 0.7 and $\lambda_2$ to 0.8. Additionally, 40 text prompts for the TE strategy were generated using ChatGPT~\cite{chatgpt}.

During testing, we apply 50-step DDIM sampling~\cite{song2020denoising} with a classifier-free guidance scale~\cite{ho2022classifier} set to 8.0.

In the experiments detailed in the main paper, we evaluated Textual Inversion on the official LDM model~\cite{rombach2022high}, while other methods, including ours, used Stable Diffusion v1.4~\cite{stablediffusionlink} as the baseline to ensure a fair comparison.

For the results presented in Table 3 of the main paper, we collected 25 images of individuals not included in the PhotoMaker~\cite{li2024photomaker} and IP-Adapter~\cite{ye2023ip} training sets. We prepared 40 text prompts and generated four images per individual for each prompt, resulting in a total of 4000 images. The evaluation was conducted using the default settings.

\subsection{User Study}
To thoroughly assess the performance of our method, we conducted a \textit{Two-Alternative Forced Choice} (2AFC) user study, focusing on pairwise comparisons from the perspective of human visual perception. Participants were presented with results from our method, PersonaMagic, alongside those from competing methods. They were asked to select the image that best matched the given prompt and the image that most closely resembled the reference. We recruited 50 participants, each of whom evaluated 120 subjects with 15 prompts per subject, as detailed in Table \ref{tab:user study}.

The results indicate that more participants recognized our method’s superior ability to preserve identity compared to Textual Inversion~\cite{gal2022image}, Custom Diffusion~\cite{kumari2023multi}, and Prefusion~\cite{tewel2023key}. While our approach marginally outperforms DreamBooth~\cite{ruiz2023dreambooth} and NeTI~\cite{alaluf2023neural} in identity preservation, it more effectively aligns with users’ imaginative interpretations of prompts. Consequently, our method demonstrates a better attunement to user preferences in customization.

\begin{table}[t]
	\centering
	\resizebox{0.47\textwidth}{!}{\begin{tabular}{l|c|c}
			\toprule
			\multirow{2}*{Method} & Text &Identity \\
			&Similarity &Preservation \\
			\midrule
			Textual Inversion~\cite{gal2022image} &  \textbf{70.02}$\pm$ 2.03 $\%$ &  \textbf{73.54}$\pm$ 1.36 $\%$ \\
			DreamBooth~\cite{ruiz2023dreambooth}        &  \textbf{78.11}$\pm$ 1.73 $\%$ &  \textbf{51.47}$\pm$ 1.67 $\%$ \\
			Custom Diffusion~\cite{kumari2023multi}  &  \textbf{52.18}$\pm$ 2.07 $\%$ &  \textbf{65.44}$\pm$ 1.82 $\%$ \\
			Perfusion~\cite{tewel2023key}         &  \textbf{59.63}$\pm$ 2.11 $\%$ &  \textbf{79.71}$\pm$ 2.03 $\%$\\
			NeTI~\cite{alaluf2023neural}         &  \textbf{82.66}$\pm$ 1.33 $\%$ &  \textbf{58.49}$\pm$ 1.90 $\%$\\
			\bottomrule
	\end{tabular}}
	\caption{User preference. Percentage of responses favoring PersonaMagic in pairwise comparisons against each competitor.}
	\label{tab:user study}
\end{table}

\begin{figure}
	\centering
	\includegraphics[width=\linewidth]{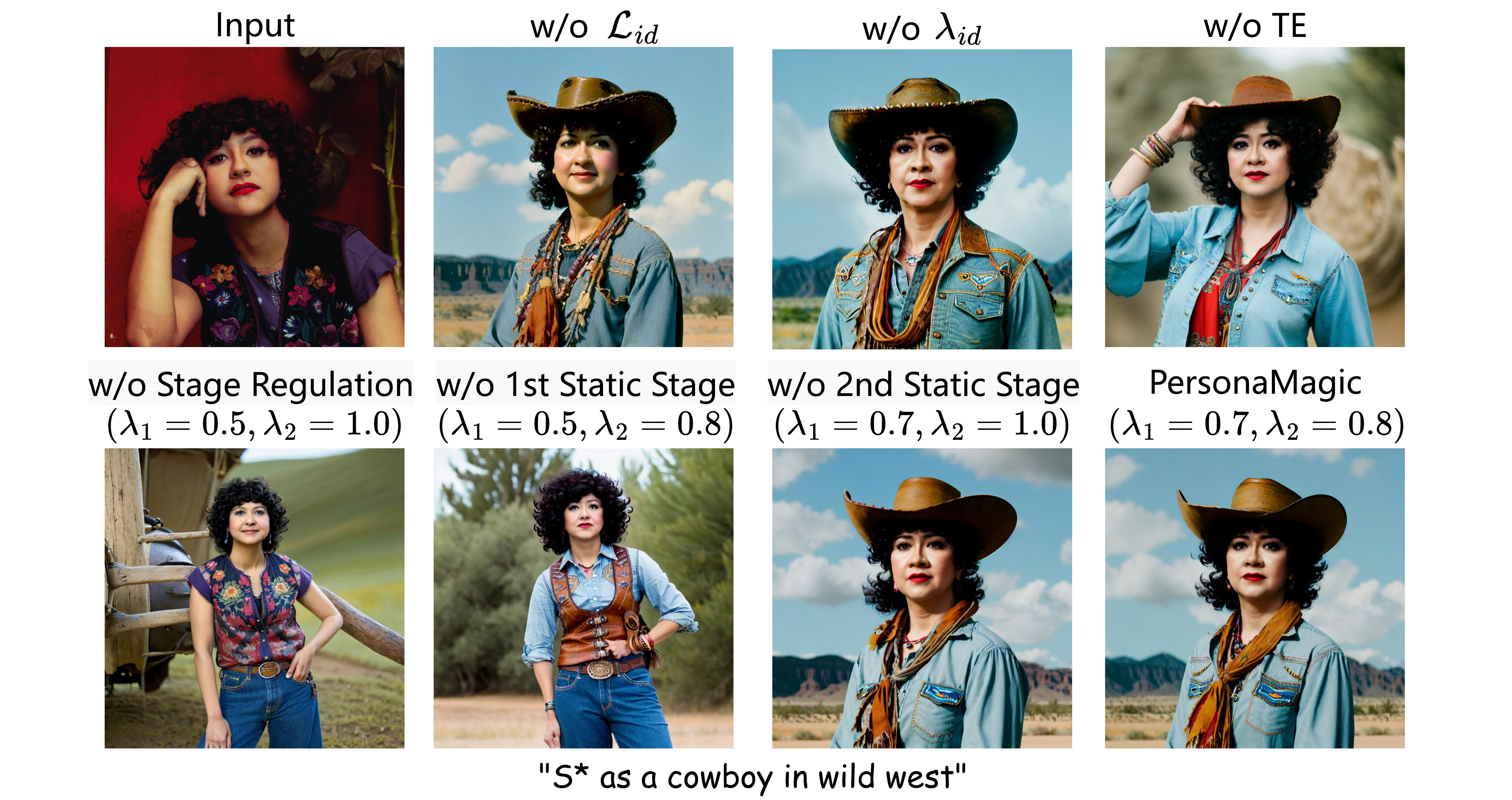}
	\caption{Qualitative ablation study of different model variants.}
	\label{fig:ablation}
\end{figure}
\begin{table}[t]
	\centering
	\setlength{\tabcolsep}{1mm}	
	\fontsize{9}{\baselineskip}\selectfont
		\begin{tabular}{l||c|c}
			\toprule
			\multirow{3}*{Method} & \multicolumn{2}{c}{CelebA-HQ} \\
			\cmidrule(lr){2-3} &Text  & Identity\\	
			&Similarity &Preservation  \\	
			\midrule
			Custom Diffusion (CD)& 0.738 & 0.283 \\	
			CD + $\lambda(t)\mathcal L_{id}$  & 0.745 & 0.270 \\
			\midrule
			PersonaMagic & \textbf{0.747} & \textbf{0.345} \\
			\bottomrule
	\end{tabular}
	\caption{The influence of identity loss $\mathcal{L}_{id}$. Identity preservation may be compromised if the temporal dynamics of diffusion models are not adequately considered.}
	\label{tab:loss}
\end{table}

\subsection{Additional Ablation Study}
To further validate the effectiveness of our approach's components, we present a qualitative ablation study, as shown in Fig.~\ref{fig:ablation}. This study examines the roles of each element, leading to the following conclusions:

1) Impact of TE Strategy Removal: When the TE strategy is removed, the generated images deviate significantly from the given text prompt. This deviation occurs because the learned embedding's high attention weight causes the diffusion model to overlook other semantic expressions.

2) Effect of Replacing $\lambda_{id}(t)$ with its Expectation: Replacing $\lambda_{id}(t)$ with its mathematical expectation $\frac{1}{T}$ over the interval $[0, T]$ results in a slight decline in identity preservation. Similar declines are observed when $\mathcal{L}_{id}$ is excluded or when the second static stage is altered ($\lambda_1=0.7$, $\lambda_2=1.0$).

3) Consequences of Omitting the First Static Stage: Without the first static stage ($\lambda_1=0.5$, $\lambda_2=0.8$), the background content in the generated images becomes suboptimal. This issue arises because the learned embedding may shift focus to areas outside the face.

4) Importance of Stage Regulation: Omitting stage regulation ($\lambda_1=0.5$, $\lambda_2=1.0$) weakens both text similarity and identity preservation. This highlights the critical role of learning new concepts during the dynamic stage.

\begin{table}[t]
	\centering
	\resizebox{0.47\textwidth}{!}{
		\begin{tabular}{l||c|c||c|c||c|c}
			\toprule
			\multirow{3}*{Method}  & \multicolumn{2}{c||}{CelebA-HQ}& \multicolumn{2}{c||}{FFHQ} &\multicolumn{2}{c}{LFW} \\
			\cmidrule(lr){2-3}
			\cmidrule(lr){4-5}		
			\cmidrule(lr){6-7}
			&Text & Identity & Text & Identity&Text & Identity\\	
			&Similarity& Preservation&Similarity& Preservation&Similarity& Preservation\\	
			\midrule
			Textual Inversion & 0.685& -0.053& 0.694& -0.023& 0.666& 0.022 \\
			DreamBooth & 0.679&\textbf{0.380}& 0.652 &\textbf{0.465}& \textbf{0.756}& 0.146 \\
			Custom Diffusion &\underline{0.738}& 0.283&\textbf{0.749}& 0.281& 0.745& 0.179 \\
			Perfusion & 0.708 & -0.073 & 0.704 & -0.027  & 0.699 & -0.006 \\		
			NeTI & 0.625 & 0.329 & 0.622 & \underline{0.425} & 0.641 & 0.227  \\		
			\midrule
			\textbf{PersonaMagic}& \textbf{0.747}& \underline{0.345}& \underline{0.747}& 0.407 & \underline{0.751} & \textbf{0.278} \\
			\bottomrule
	\end{tabular}}
	\caption{Quantitative evaluation values based on Stable Diffusion v1.4, with Textual Inversion results from LDM due to its superior performance compared to the Stable Diffusion model.}
	\label{tab:Quantitative Evaluation}
\end{table}

On the other hand, while it might seem intuitive that identity loss $\mathcal{L}_{id}$ could provide more facial identity information for embedding learning, its actual impact on improving identity accuracy is less straightforward. To investigate this, we integrated identity loss into Custom Diffusion and trained the model on CelebA-HQ, with the results summarized in Table~\ref{tab:loss}. Contrary to expectations, the results showed not only a lack of improvement in identity metrics but also a slight decline.

Our analysis revealed that at larger timesteps, the prediction $X_{0|t}$ from the noisy image $X_t$ becomes too blurred, making it difficult for identity loss to provide accurate guidance. Even with timestep-specific weight adjustments using $\lambda(t)$, the errors caused by information loss at larger timesteps persisted. To address these challenges and improve identity accuracy, we employed a stage-regulation strategy, which proved more effective in mitigating these issues.

\begin{figure}
	\centering
	\includegraphics[width=\linewidth]{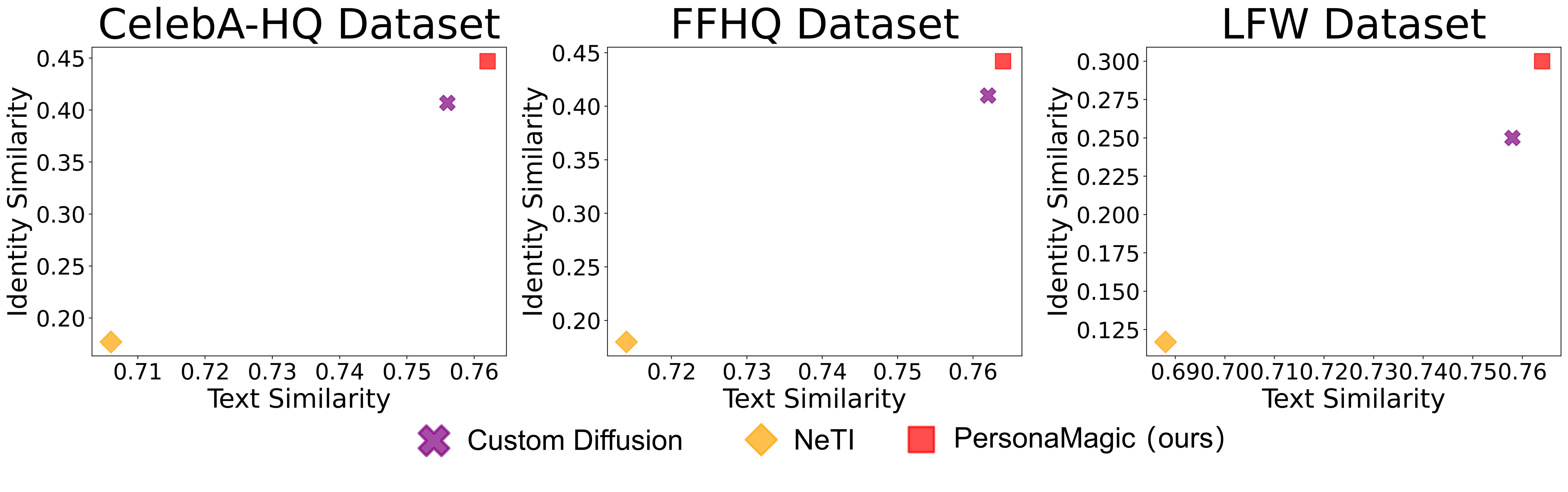}
	\caption{Quantitative evaluation based on Dreamlike Photoreal v2.0 demonstrates the robustness of PersonaMagic.}
	\label{fig:dreamlike}
\end{figure}

\subsection{Quantitative Evaluation}
First, to concretely demonstrate the quantitative performance of our method, we present the metric values for the comparisons in Table~\ref{tab:Quantitative Evaluation}, corresponding to Fig. 7 in the main paper. To further showcase the robustness of our method across different pretrained models, we conducted an additional quantitative evaluation using Dreamlike Photoreal v2.0~\cite{dreamlikelink} as the base text-to-image diffusion model, in addition to the results on Stable Diffusion v1.4. These results are illustrated in Fig.~\ref{fig:dreamlike}, with detailed values provided in Table~\ref{tab:Quantitative dreamlike}.

Notably, we included only Custom Diffusion and NeTI in this experiment, as other competitors, such as Textual Inversion and Perfusion, are incompatible with the Dreamlike model. Additionally, training DreamBooth would exceed the memory limits of an RTX 3090 GPU. As shown in the results, our method does not rely on a specific pretrained model and consistently outperforms all competitors across different models, achieving the best fidelity-editability balance by sitting on the Pareto front.

\begin{table}[t]
	\centering
	\resizebox{0.47\textwidth}{!}{
		\begin{tabular}{l||c|c||c|c||c|c}
			\toprule
			\multirow{3}*{Method}  & \multicolumn{2}{c||}{CelebA-HQ}& \multicolumn{2}{c||}{FFHQ} &\multicolumn{2}{c}{LFW} \\
			\cmidrule(lr){2-3}
			\cmidrule(lr){4-5}		
			\cmidrule(lr){6-7}
			&Text & Identity & Text & Identity&Text & Identity\\	
			&Similarity& Preservation&Similarity& Preservation&Similarity& Preservation\\	
			\midrule
			Custom Diffusion &0.756& 0.407&0.762& 0.410& 0.758& 0.250 \\
			NeTI & 0.706 & 0.177 & 0.714 & 0.180 & 0.688& 0.117 \\		
			\midrule
			\textbf{PersonaMagic}& 0.762& 0.447& 0.764& 0.442& 0.764 & 0.300 \\
			\bottomrule
	\end{tabular}}
	\caption{Quantitative evaluation values based on Dreamlike Photoreal v2.0.}
	\label{tab:Quantitative dreamlike}
\end{table}

\subsection{Applications}

\begin{figure}
	\centering
	\includegraphics[width=\linewidth]{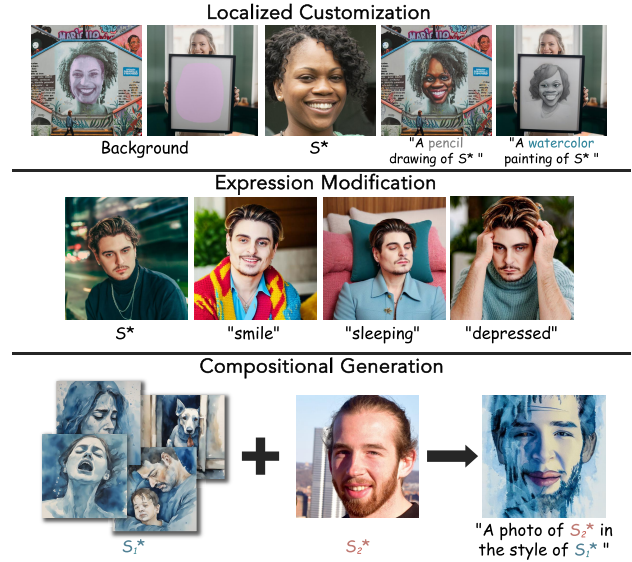}
	\caption{Our method can be applied to various downstream tasks. From top to bottom: Localized Customlization, Expression Modification, and Compositional Generation. }
	\label{fig:applications}
\end{figure}

We apply our method to several downstream applications to demonstrate its flexibility and applicability.

\noindent\textbf{Localized Customlization.}
Real-world backgrounds can provide users with ample inspiration. We combine our method with Blended Latent Diffusion~\cite{avrahami2023blended} for localized customization, as shown in upper panel of Fig. \ref{fig:applications}. Users manually draw an arbitrarily shaped mask on a background, and then our framework paints the desired foreground character in the mask area according to another face image. The customization results retain the appearance of the reference character while incorporating the background with the assistance of prompts.

\noindent\textbf{Expression Manipulation.}
Our method enables expression manipulation guided by text prompts without compromising facial fidelity. The advantage of employing a customization approach for editing lies in its capacity to seamlessly integrate individuals into new scenes based on facial expressions.
As illustrated in the middle panel of Fig.~\ref{fig:applications}, using ``sleeping" for editing results in the person closing their eyes and assuming a lying pose.

\noindent\textbf{Compositional Generation.}
Text prompts allow for the natural combination of different concepts. We employed the prompts ``A painting in the style of $S_1*$" and ``A photo of $S_2*$" to learn the artistic style of a set of images and a facial image, respectively. During inference, we utilized the prompt ``A photo of $S_2*$ in the style of $S_1*$" to transfer styles, as illustrated in the lower panel of Fig.~\ref{fig:applications}.
\begin{figure}
	\centering
	\includegraphics[width=\linewidth]{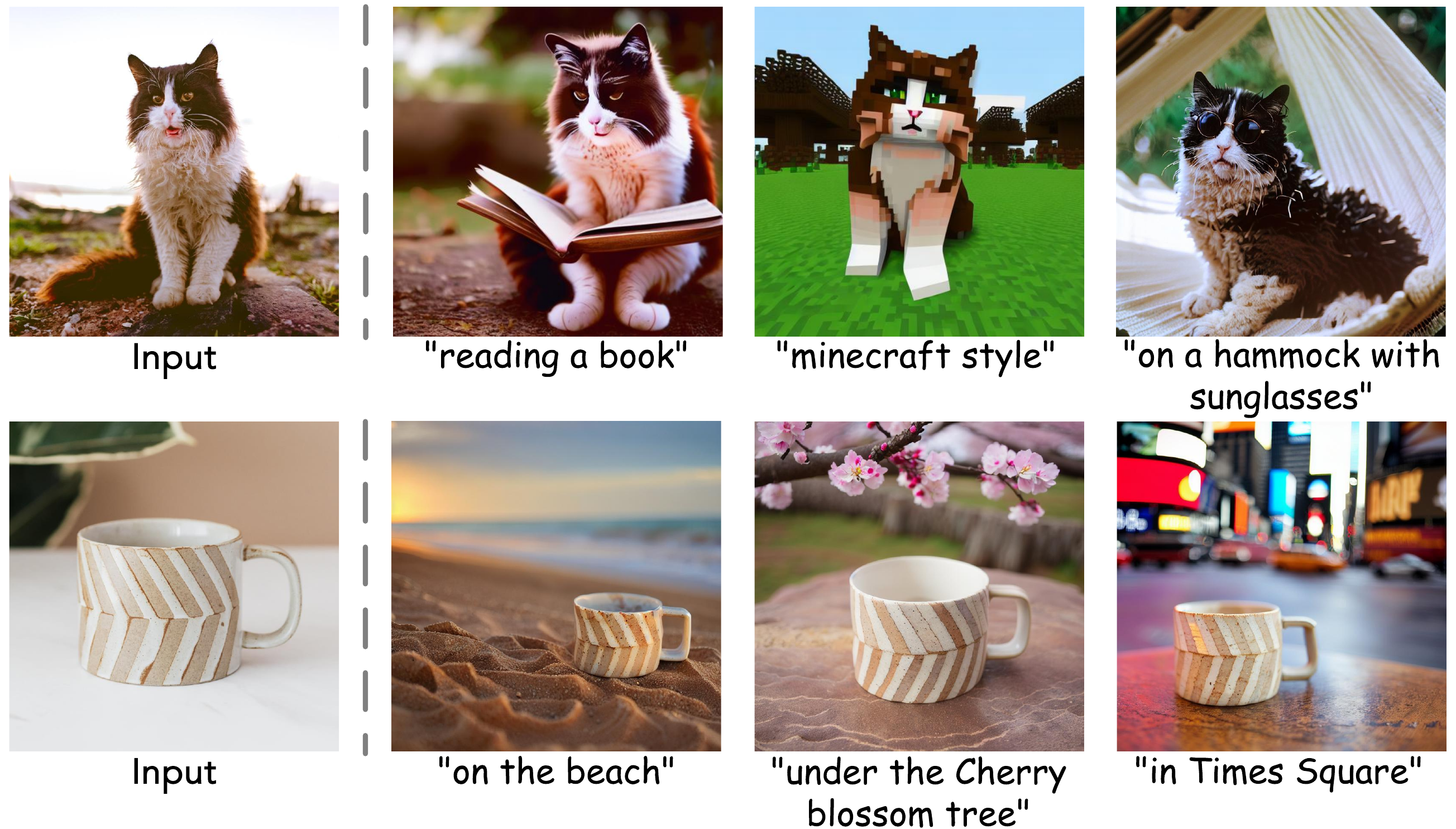}
	\caption{PersonaMagic can adapt to non-facial domains, showcasing its generality beyond facial content. }
	\label{fig:other domains}
\end{figure}

\noindent\textbf{Application in Other Domains.}
Our method is not only competent for face customization but can also be applied in other domains, such as animals and man-made objects, as illustrated in Fig.~\ref{fig:other domains}. Even without $\mathcal{L}_{id}$, the outcomes maintain strong identity preservation, particularly evident in the cat's fur and the cup's texture, which align well with the given image.
This is attributed to our stage regulation strategy, making it easier to represent more finer details of the target object across different timesteps.

\subsection{Limitations}
\begin{figure}
	\centering
	\includegraphics[width=\linewidth]{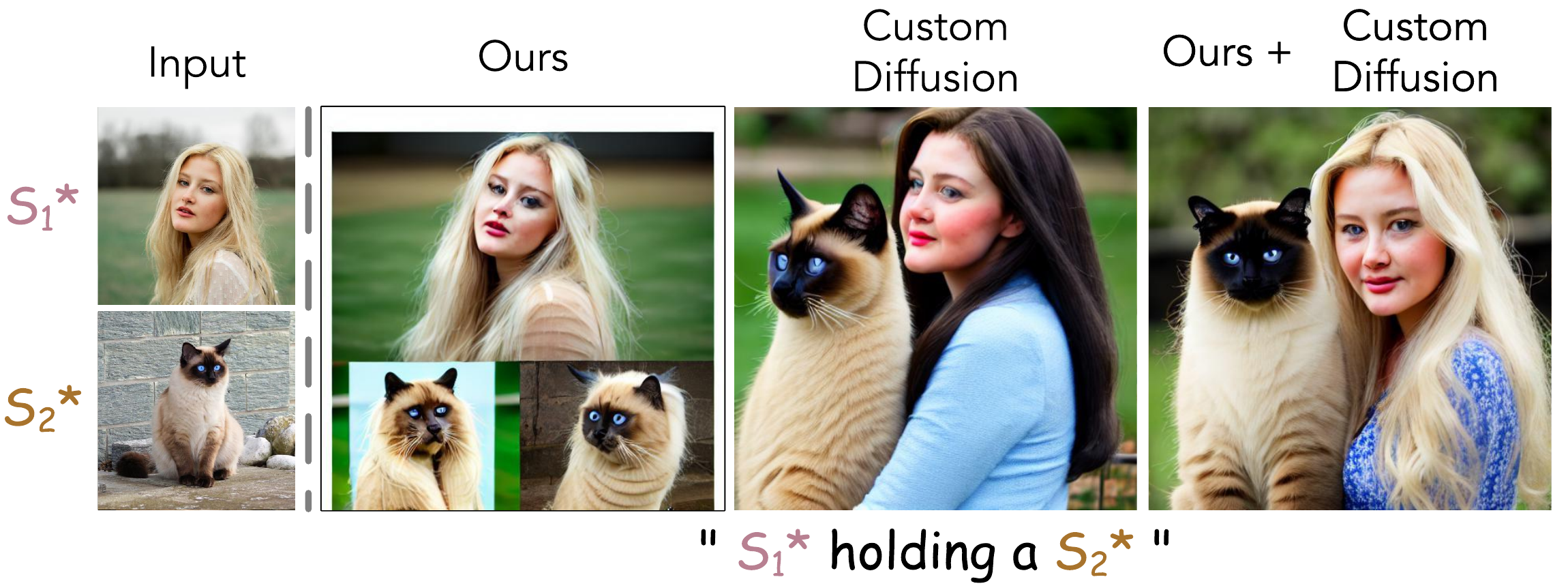}
	\caption{Failure cases of PersonaMagic.}
	\label{fig:failure}
\end{figure}

While our method has shown effectiveness in single-face customization, it faces challenges when attempting to combine multiple concepts, such as generating both a specific person and a particular cat in the same image. This limitation arises from the absence of fine-tuning for the query weights in the cross-attention layers. However, by adapting our stage-regulated embeddings to a fine-tuning-based approach, this issue can be effectively addressed. As illustrated in Fig.~\ref{fig:failure}, integrating our method as a plug-in to Custom Diffusion enables successful multi-concept personalization, even in one-shot scenarios.

\begin{figure}
	\centering
	\includegraphics[width=\linewidth]{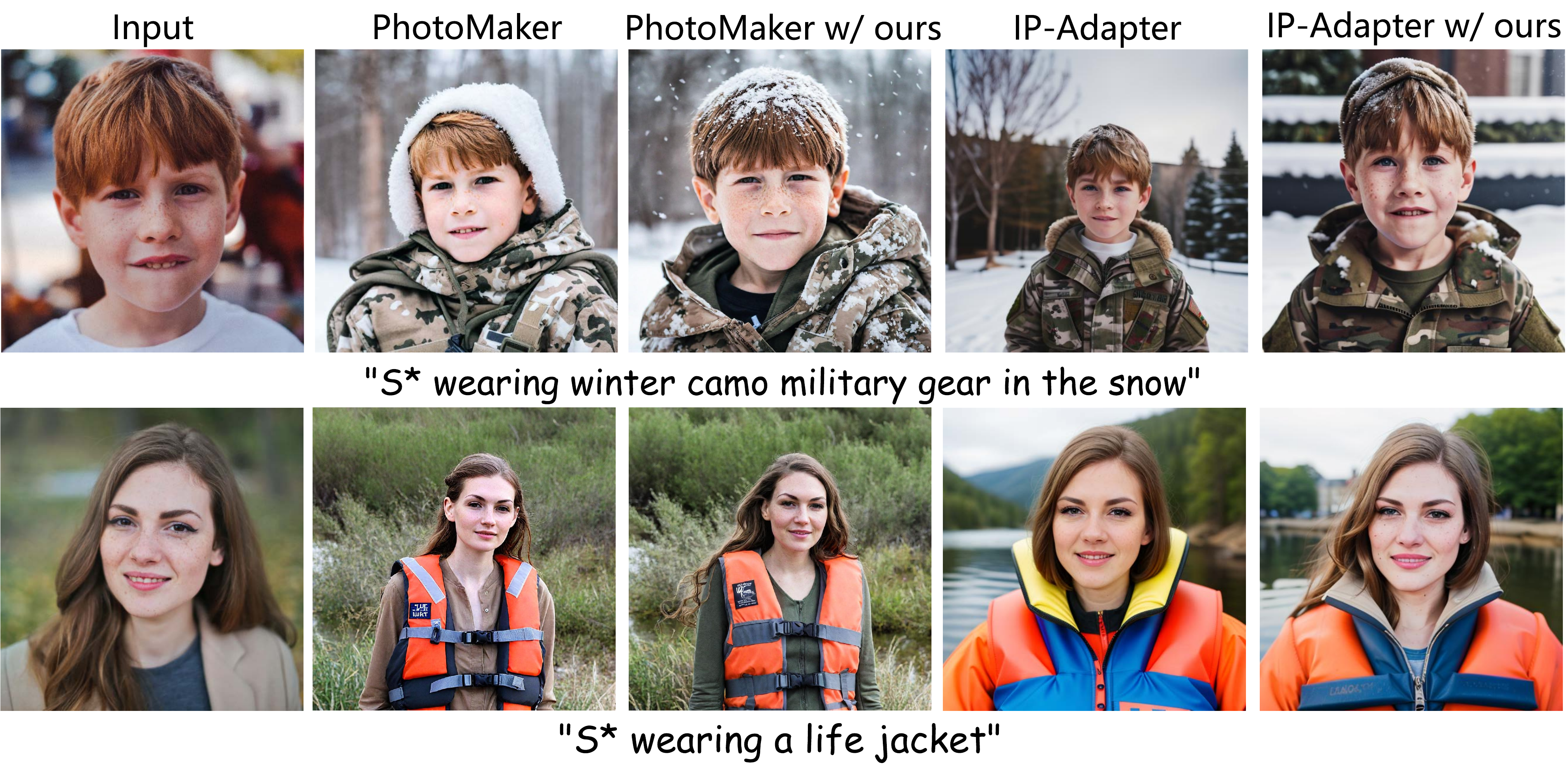}
	\caption{Integrating PersonaMagic into pre-trained personalization models refines facial details in results. } 
	\label{fig:plugin}
\end{figure}
\subsection{Additional Visual Results}
We present additional visual results of PersonaMagic with various prompts in Fig.~\ref{fig:add}. As shown, PersonaMagic successfully personalizes individuals with accurate identity across a range of text prompts, including those specifying clothing, actions, styles, or backgrounds.

To further demonstrate the flexibility of our method, we provide visual comparisons before and after integrating it with the pre-trained personalization models PhotoMaker~\cite{li2024photomaker} and IP-Adapter~\cite{ye2023ip}, as depicted in Fig.~\ref{fig:plugin}, Fig.~\ref{fig:photomaker}, and Fig.~\ref{fig:ip-adapter}, respectively.

In Fig.~\ref{fig:plugin}, both Photomaker and IP-Adapter exhibit limitations in identity preservation when handling unseen facial concepts. After integrating our method, the generated faces preserved finer details, such as freckles, which were missing in the baseline results.

For PhotoMaker, as shown in Fig.~\ref{fig:photomaker}, the model struggled to accurately restore the subject's beard and hairstyle. However, with our stage-regulated embedding, these identity features were preserved, resulting in significantly improved identity consistency.

Similarly, as illustrated in Fig.~\ref{fig:ip-adapter}, IP-Adapter initially failed to match the facial shape of the given subjects. After integrating our method, identity accuracy was improved, and alignment with the user-provided prompt was maintained.

\begin{figure*}
	\centering
	\includegraphics[width=\linewidth]{AAAI_figure/ours_more.pdf}
	\caption{More visual results of PersonaMagic on celebrities and non-celebrities, along with evaluation prompts, demonstrate our method's excellence in both high identity preservation and text alignment.}
	\label{fig:add}
\end{figure*}

\begin{figure*}
	\centering
	\includegraphics[width=\linewidth]{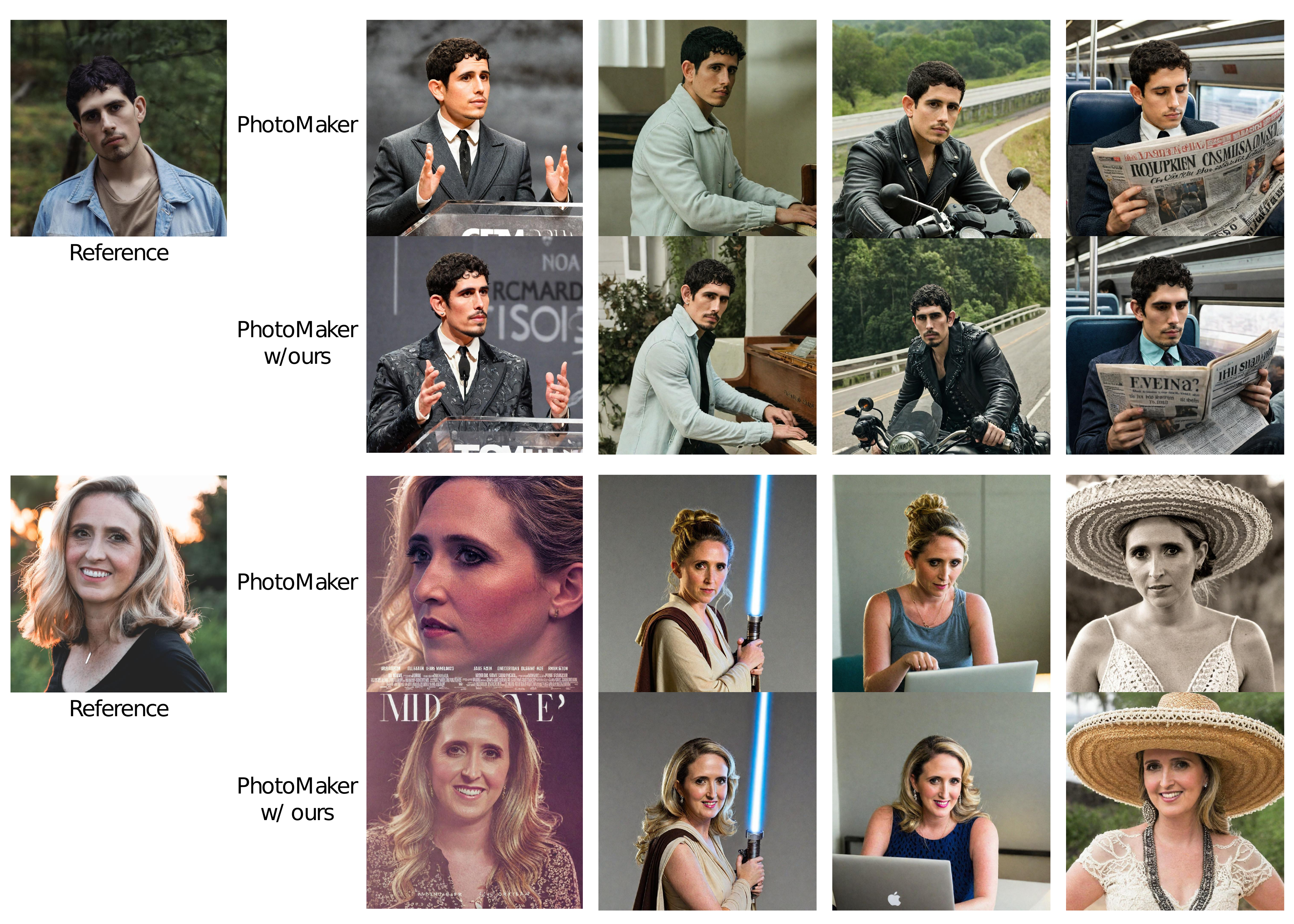}
	\caption{Additional visual results of PersonaMagic integrated as a plug-in for pre-trained personalization model PhotoMaker.}
	\label{fig:photomaker}
\end{figure*}

\begin{figure*}
	\centering
	\includegraphics[width=\linewidth]{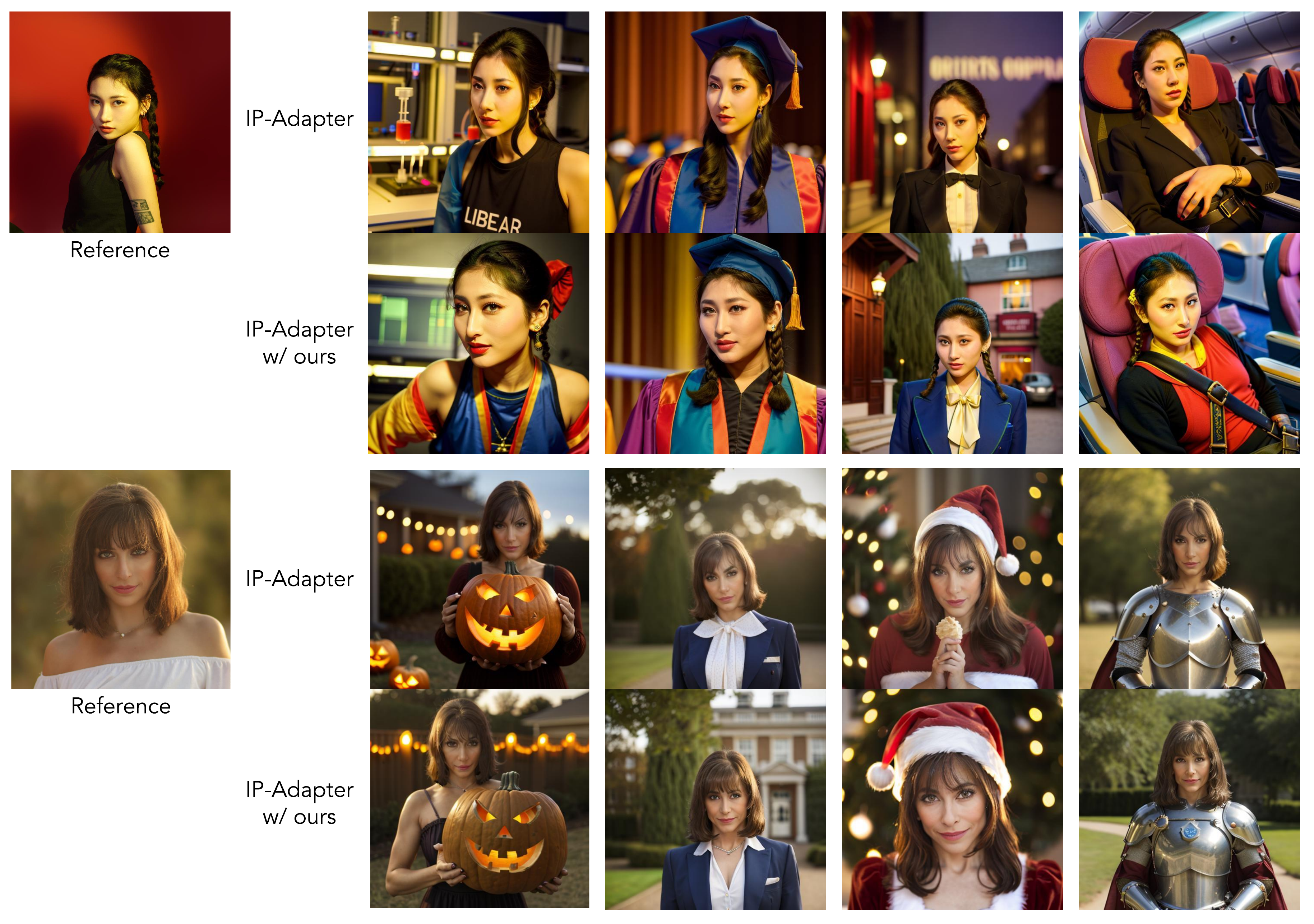}
	\caption{Additional visual results of PersonaMagic integrated as a plug-in for pre-trained personalization model IP-Adapter.}
	\label{fig:ip-adapter}
\end{figure*}

\bibliography{ref}

\end{document}